\documentclass[runningheads]{llncs}
\usepackage{graphicx}
\usepackage{amsmath}
\usepackage{subcaption}
\usepackage{rotating}
\usepackage{multirow}
\usepackage{pgfplots}
\pgfplotsset{compat=newest}
\usepackage{cite}
\usepackage{comment}

\usepackage{xcolor, soul}

\begin{document}
\title{
FastHebb: Scaling Hebbian Training of Deep Neural Networks to ImageNet Level
\thanks{This work was partially supported by the H2020 project AI4Media (GA 951911).}}
\titlerunning{FastHebb}
%
\author{Gabriele Lagani\inst{1,2} \and
Claudio Gennaro\inst{2} \and
Hannes Fassold\inst{3} \and
Giuseppe Amato\inst{2}}
\authorrunning{G. Lagani et al.}
%
\institute{Dept. of Computer Science, University of Pisa, 56127, Pisa, Italy \\ \email{gabriele.lagani@phd.unipi.it} \and
ISTI-CNR, 56124, Pisa, Italy \\
\email{\{gabriele.lagani,giuseppe.amato,claudio.gennaro\}@isti.cnr.it} \and
Joanneum Research, 8010, Graz, Austria \\
\email{hannes.fassold@joanneum.at}}
\maketitle              
\begin{abstract}
Learning algorithms for Deep Neural Networks are typically based on supervised end-to-end Stochastic Gradient Descent (SGD) training with error backpropagation (backprop). Backprop algorithms require a large number of labelled training samples to achieve high performance. However, in many realistic applications, even if there is plenty of image samples, very few of them are labelled, and semi-supervised sample-efficient training strategies have to be used. Hebbian learning represents a possible approach towards sample efficient training; however, in current solutions, it does not scale well to large datasets. 
In this paper, we present \textit{FastHebb}, an efficient and scalable solution for Hebbian learning which achieves higher efficiency by 1) merging together update computation and aggregation over a batch of inputs, and 2) leveraging efficient matrix multiplication algorithms on GPU.
We validate our approach on different computer vision benchmarks, in a semi-supervised learning scenario. FastHebb outperforms previous solutions by up to 50 times in terms of training speed, and notably, for the first time, we are able to bring Hebbian algorithms to ImageNet scale.

\end{abstract}

\section{Introduction}

In the past few years, Deep Neural Networks (DNNs) have emerged as a powerful technology in the domain of computer vision \cite{krizhevsky2012, he2016}. DNNs started gaining popularity also in the domain of large scale multimedia Content-Based Image Retrieval (CBIR), replacing handcrafted feature extractors \cite{wan2014, babenko2014} and using activations of internal layers as feature vectors for similarity search. Learning algorithms for DNNs are typically based on supervised end-to-end Stochastic Gradient Descent (SGD) training with error backpropagation (\textit{backprop}). This approach is considered biologically implausible by neuroscientists \cite{oreilly}. Instead, they propose \textit{Hebbian} learning as a biological alternative to model synaptic plasticity \cite{haykin}. 

Moreover, backprop-based algorithms need a large number of labeled training samples in order to achieve high results, which are expensive to gather, as opposed to unlabeled samples. Therefore, researchers started to investigate \textit{semi-supervised} learning strategies, which aim to exploit large amounts of unlabeled data, in addition to the fewer labeled data, for sample efficient learning \cite{bengio2007, larochelle2009}. 
In this context, a possible direction that has been proposed is to perform an unsupervised pre-training stage on all the available samples, which is then followed by a supervised fine-tuning stage on the few labeled samples only \cite{kingma2014b, zhang2016}.

In recent work, Hebbian learning has begun to gain attention from the computer science community as an effective method for unsupervised pre-training, since Hebbian algorithms do not require supervision, achieving promising results in scenarios with scarce labeled data \cite{lagani2021b, lagani2022b}.
However, current solutions for Hebbian training (such as \cite{lagani2021c, lagani2021d, lagani2022a, wadhwa2016b, bahroun2017, krotov2019}) are still limited in terms of computational efficiency, making it difficult to scale to large datasets such as ImageNet \cite{imagenet}.

In order to address this issue, we present \textit{FastHebb}, a novel solution for Hebbian training that achieves enhanced efficiency by leveraging two observations. First, when a mini-batch of inputs has to be processed, the weight update corresponding to each input is first computed, and then the various updates are aggregated over the mini-batch; however, update computation and aggregation can be merged together with a significant speedup. Second, Hebbian learning rules can be reformulated in terms of matrix multiplications, which enables to exploit efficient matrix multiplication algorithms on GPU.

We validate our method on various computer vision benchmarks. Since Hebbian algorithms are unsupervised, we consider a semi-supervised training scenario, in which Hebbian learning is used to perform unsupervised network pre-training, followed by fine-tuning with traditional backprop-based supervised learning. We also consider sample efficiency scenarios, in which we assume that only a small fraction of the training data is labeled, in order to study the effectiveness of Hebbian pre-training in scenarios with scarce data. In order to make comparisons with backprop-based methods, we consider Variational Auto-Encoder (VAE) \cite{kingma2013, kingma2014b} pre-training as a baseline for comparisons. We show that our approach achieves comparable results, but with a significant speed-up both in terms of number of epochs, as well as total training time. In particular, our method achieves up to 50x speed-up w.r.t. previous Hebbian learning solutions, allowing to scale up our experiments to ImageNet level. To the best of our knowledge, this is the first time that Hebbian algorithms are applied at such scale. 

In summary, our contribution is twofold:
\begin{enumerate}
    \item We propose a novel efficient solution to Hebbian learning algorithms, with code available online \footnote{github.com/GabrieleLagani/HebbianLearning/tree/fasthebb};
    \item We performed extensive experimental evaluation of the performance of our solution on various computer vision benchmarks. In particular, for the first time (to the best of our knowledge) results of Hebbian algorithms on ImageNet are provided.
\end{enumerate}

The remainder of this paper is structured as follows:
Section \ref{sec:rel_work} introduces some background and related work on Hebbian learning;
Section \ref{sec:fasthebb} presents our FastHebb method;
Section \ref{sec:experiments} provides the details of our experimental setup;
Section \ref{sec:results} presents the results of our experiments;
Finally, Section \ref{sec:conclusions} outlines some concluding remarks and hints for future work.

\section{Background and related work} \label{sec:rel_work}

In this section, we illustrate some of the Hebbian learning rules from literature that recently provided promising results, and we describe some related work focusing on the application of such rules on computer vision tasks, in particular in semi-supervised training scenarios. Since a thorough explanation of the Hebbian rules would be outside the scope of this paper, here we just give the update equations of interest, referring the interested reader to the vast literature on the topic \cite{haykin, gerstner, lagani2022a, lagani2021a, lagani2021d}.

Let us start by considering a neuron, identified by an index $i$, with weight vector $\mathbf{w}_i$, which receives as input a vector $\mathbf{x}$, and produces a corresponding output $y_i$.
One of the Hebbian approaches that we focus on is the \textit{soft Winner-Takes-All} (SWTA) competitive learning rule \cite{grossberg1976a, nowlan1990, lagani2021d}, which can be expressed as follows:
\begin{equation} \label{eq:hwta}
    \Delta \mathbf{w}_i = \eta \, r_i \, (\mathbf{x} - \mathbf{w}_i)
\end{equation}
where $\eta$ is the learning rate, and the coefficient $r_i$ is a score computed as the \textit{softmax} of the neural activations: $r_i = \frac{e^{y_i/T}}{\sum_j e^{y_j/T}}$. 
Here, T is the \textit{temperature} parameter of the softmax, which serves to cope with the variance of the activations (the name comes from statistical mechanics, where this operation was first defined). The effect of such a defined score is to allow each neuron to specialize on a different cluster of input patterns.

The other learning rule that we consider is Hebbian Principal Component Analysis (HPCA) \cite{karhunen1995, becker1996a, lagani2021b, lagani2022a}:
\begin{equation} \label{eq:hpca}
    \Delta \mathbf{w}_i = \eta \, y_i \, \Big( \mathbf{x} - \sum_{j=1}^i y_j \mathbf{w_j} \Big)
\end{equation}

WTA competition was studied in past work as a possible approach for training relatively shallow neural networks \cite{wadhwa2016b, krotov2019} (with up to 2-3 hidden layers). The investigation was further extended to deeper networks, and to hybrid architectures where some layers were trained by backprop and others by Hebbian learning \cite{lagani2019, lagani2022a}.
Experimental results on CNNs showed promises of HPCA-like learning mechanisms initially with shallow networks \cite{bahroun2017}, and then with deeper networks as well \cite{lagani2021b, lagani2021c, lagani2022a}.

Since the HPCA and SWTA learning rules are unsupervised, they have found application in the context of semi-supervised neural network training, in order to perform an unsupervised pre-training stage \cite{lagani2021b, lagani2021c, lagani2021d, lagani2022b}. In particular, they were found to be particularly useful in sample efficient learning scenarios, i.e. situations with scarce availability of labeled data. Related approaches for unsupervised pre-training are based on autoencoding architectures \cite{bengio2007, larochelle2009, kingma2014b, zhang2016}. Results on various computer vision benchmarks suggest that Hebbian pre-training allows to significantly improve performance on such scenarios compared to other unsupervised pre-training methods such as Variational Auto-Encoder (VAE) pre-training \cite{kingma2013, kingma2014b}. Application of Hebbian learning to semi-supervised settings seems a promising direction. Other approaches to semi-supervised learning are based on pseudo-labeling/consistency-based methods \cite{iscen2019, sellars2021}. However, these methods are not in contrast with unsupervised pre-training, and they could actually be integrated together. This possible future direction will also be highlighted in Section \ref{sec:conclusions}.

The problem with current Hebbian learning solutions is that they do not scale well to large datasets. Note that, the learning rules mentioned above describe the weight update for a single input $\mathbf{x}$. When there is a batch of inputs to be processed, the weight updates are aggregated over the batch dimension, typically by averaging (or weighted averaging, for SWTA, the weights being the competition scores $r_i$, check \cite{lagani2019, thesis} for details). Similarly, in a convolutional layer, $\mathbf{x}$ would correspond to a patch extracted from an input at a given offset, and weight updates computed at different offsets need to be aggregated over all the extracted patches. In this contribution, we notice that these two phases (update computation and aggregation) can be merged together, which allows to reformulate Hebbian learning rules more efficiently in terms of matrix multiplications, which are particularly suitable for GPU computation. We show that our solution is able to scale well to large datasets such as ImageNet.

\section{Efficient Hebbian learning with FastHebb method} \label{sec:fasthebb}

Let us start by introducing some preliminary information about the multi-dimensional tensor data that we need to work with, and the notation that will be used in the following.

\begin{figure}
    \centering
    \includegraphics[width=0.95\textwidth]{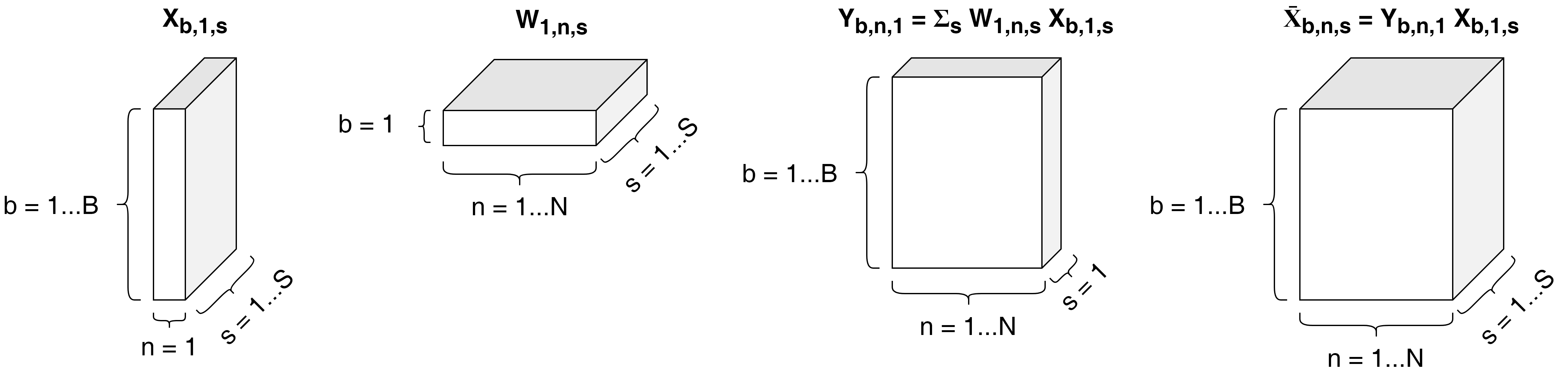}
    \caption{Types of tensor objects involved in our scenario.}
    \label{fig:tensors}
\end{figure}

We define a \textit{tensor} simply as a multi-dimensional array of data. In particular, our tensors are three-dimensional. We denote such tensors with capital letters, followed by as many indices as dimensions (three in our case). A dimension of size 1, also known as a \textit{singleton} dimension, is denoted with the symbol 1 as index. Moreover, we adopt the following convention: index $b=1...B$ denotes the \textit{batch} dimension, index $n=1...N$ denotes the \textit{neuron} dimension, and index $s=1...S$ denotes the \textit{size} dimension. Note that the meaning of an index is inferred by the corresponding letter and not by its position. With reference to Fig. \ref{fig:tensors}, the first tensor (from left to right) is a typical input tensor, consisting of a mini-batch of $B$ inputs, each being a vector of size $S$. The second tensor represents a typical weight matrix, consisting of one weight vector for each of the $N$ neurons, each of size $S$. The third is a typical output tensor, with each output being a vector on $N$ elements, one for each neuron, and there is one such vector for each of the $B$ elements in the batch. The last is a typical reconstruction tensor, which extends over all the dimensions.

Finally, in order to make the use of matrix multiplication explicit in our formulas, we will use the notation $\textrm{matmul}(\cdot, \cdot)$ as follows:
\begin{equation}
    C_{d, e, g} = \sum_f A_{d, e, f} B_{d, g, f} = \sum_f A_{d, e, f} B_{d, f, g} := \textrm{matmul}(A_{d, e, f}, B_{d, f, g})
\end{equation}
Note that we are taking the tensor product between tensors $A$ and $B$, identifying index $d$ and contracting index $f$. This corresponds to a batch matrix multiplication over index $d$, i.e. mapping $d$ pairs of matrices with indices $(e, f)$ and $(f, g)$, to $d$ matrices with indices $(e, g)$: $(e, f) \times (f, g) \rightarrow (e, g) $. If more that three dimensions are present, then the last two denote height and width of the matrices, and all the previous dimensions are considered as batch dimensions (and thus identified). If a batch dimension of one of the multiplied tensors happens to be a singleton, then it undergoes \textit{broadcasting} to match the other tensor dimension, as done in common mathematical frameworks. In all the other cases the corresponding batch dimensions of the two tensors must have the same size (as well as the contracted $f$ dimension). Sums, subtractions, and multiplications by constants over tensor are performed component-wise, but all dimensions must match. Also in this case, a singleton dimension of one tensor undergoes broadcast to match the corresponding dimension of the other tensor (in case of singleton dimensions, and only in this case, correspondence is inferred from the position of the indices).


Using the notation introduced above, we can express the Hebbian rules discussed in this paper, including the aggregation step, as follows:

\begin{equation}
    \Delta W_{1, n, s} = \sum_b C_{b, n, 1} \, \Delta W_{b, n, s} = \textrm{matmul}(C_{n, 1, b}, \Delta W_{n, b, s})
\end{equation}

Tensor C represents the coefficients for (weighted) averaging during the update aggregation step. With our notation, we consider the batch index $b$ to run over all the patches extracted from the inputs, and also over all the inputs in the mini-batch. In other words, all the patches extracted from all the images in the mini-batch are considered as a unique larger mini-batch over which aggregation is performed.

Notice that, at this point, update computation and aggregation phases are considered together. In fact, merging these two phases is an essential step towards achieving the performance improvement addressed in this work, as described below. In particular, as the dimension associated with index $b$ is very large, since it runs over all the patches extracted from all the inputs, it would be beneficial to contract this index as soon as possible in our computations, possibly before larger tensors such as $\Delta W_{b, n, s}$ are obtained. We proceed differently depending on the Hebbian rule under consideration.

\paragraph{Hebbian Winner-Takes-All.}

The (soft-)WTA learning rule can be rewritten as follows:
\begin{equation}
\begin{split}
    \Delta W_{1, n, s} & = \eta \, \sum_b C_{b, n, 1} \, R_{b, n, 1} \, \Big ( X_{b, 1, s} - W_{1, n, s} \Big ) \\
    & = \eta \, \sum_b ( C \, R )_{b, n, 1} \, ( X - W )_{b, n, s} \\
    & = \eta \, \textrm{matmul}\Big ( ( C \, R )_{n, 1, b}, (X - W )_{n, b, s} \Big )
\end{split}
\end{equation}
Where $C_{b, n, 1} = \frac{R_{b, n, 1}}{\sum_b R_{b, n, 1}}$.

Note that this formulation requires $O(B \, N \, S)$ complexity both in time and space. In particular, it needs to store a $B \times N \times S$ tensor. All the elements are stored simultaneously in order parallelize operations over each dimension through vectorized or GPU hardware. If the amount of memory required is prohibitive, it is possible to serialize computations over one or more dimensions. However, computational performance can be improved by rewriting:
\begin{equation}
\begin{split}
    \Delta W_{1, n, s} & = \eta \, \sum_b C_{b, n, 1} \, R_{b, n, 1} \, \Big ( X_{b, 1, s} - W_{1, n, s} \Big ) = \\
    & = \eta \, \sum_b ( C \, R )_{b, n, 1} \, X_{b, 1, s} - \eta \, \sum_b ( C \, R )_{b, n, 1} \, W_{1, n, s} = \\
    & = \eta \, \textrm{matmul}\Big ( ( C \, R )_{1, n, b}, X_{1, b, s} \Big ) - \eta \, \sum_b ( C \, R )_{b, n, 1} \, W_{1, n, s} = \\
    & = \eta \, \textrm{matmul}\Big ( ( C \, R )_{1, n, b}, X_{1, b, s} \Big ) - \eta \, Q_{1, n, 1} \, W_{1, n, s}
\end{split}
\end{equation}
Where $Q_{1, n, 1} = \sum_b ( C \, R )_{b, n, 1}$.

By contracting index $b$ early, we have obtained a new formulation that requires only $O(N(B + S))$ space. The time complexity depends on the algorithm employed for matrix multiplication, which can be made lower than $O(BNS)$.

\paragraph{Hebbian Principal Component Analysis.}

The Hebbian PCA learning rule can be rewritten as follows:
\begin{equation}
\begin{split}
    \Delta W_{1, n, s} & = \eta \, \frac{1}{B} \sum_b Y_{b, n, 1} \, \Big ( X_{b, 1, s} - \sum_{n'=1}^n Y_{b, n', 1} \, W_{1, n', s} \Big ) \\
    & = \eta \, \frac{1}{B} \sum_b Y_{b, n, 1} \, \Big ( X_{b, 1, s} - \sum_{n'=1}^N L_{n, n'} \, Y_{b, n', 1} \, W_{1, n', s} \Big ) \\
    & = \eta \, \frac{1}{B} \sum_b Y_{b, n, 1} \, E_{b, n, s} \\
    & = \eta \, \frac{1}{B} \textrm{matmul} \Big ( Y_{n, 1, b}, E_{n, b, s} \Big )
\end{split}
\end{equation}
Where $E_{b, n, s} = \Big ( X_{b, 1, s} - \sum_{n'=1}^N L_{n, n'} \, Y_{b, n', 1} \, W_{1, n', s} \Big ) $, and $L_{n, n'}$ is simply a lower-triangular matrix with all ones on and below the main diagonal and all zeros above.

In this case, the computation requires $O(B N^2 S)$ space and time, but this can be improved by rewriting:

\begin{equation}
\begin{split}
    \Delta W_{1, n, s} & = \eta \, \frac{1}{B} \sum_b Y_{b, n, 1} \, \Big ( X_{b, 1, s} - \sum_{n'=1}^N L_{n, n'} \, Y_{b, n', 1} \, W_{1, n', s} \Big ) \\
    & = \eta \, \frac{1}{B} \sum_b Y_{b, n, 1} \, X_{b, 1, s} - \eta \, \frac{1}{B} \sum_b Y_{b, n, 1} \, \sum_{n'=1}^N L_{n, n'} \, Y_{b, n', 1} \, W_{1, n', s} \\
    & = \eta \, \frac{1}{B} \textrm{matmul} \Big ( Y_{1, n, b}, X_{1, b, s} \Big ) - \eta \, \frac{1}{B} \sum_{n'=1}^N \sum_b Y_{b, n, 1} \, Y_{b, n', 1} \, L_{n, n'} \, W_{1, n', s} \\
    & = \eta \, \frac{1}{B} \textrm{matmul} \Big ( Y_{1, n, b}, X_{1, b, s} \Big ) - \eta \, \frac{1}{B} \sum_{n'=1}^N \textrm{matmul} \Big ( Y_{1, n, b}, Y_{1, b, n'} \Big ) \, L_{n, n'} \, W_{1, n', s} \\
    & = \eta \, \frac{1}{B} \textrm{matmul} \Big ( Y_{1, n, b}, X_{1, b, s} \Big ) - \eta \, \frac{1}{B} \sum_{n'=1}^N P_{1, n, n'} \, W_{1, n', s} \\
    & = \eta \, \frac{1}{B} \textrm{matmul} \Big ( Y_{1, n, b}, X_{1, b, s} \Big ) - \eta \, \frac{1}{B} \textrm{matmul} \Big ( P_{1, n, n'} , W_{1, n', s} \Big ) \\
\end{split}
\end{equation}
Here, $P_{1, n, n'} = \textrm{matmul} \Big ( Y_{1, n, b}, Y_{1, b, n'} \Big ) \, L_{n, n'}$.

This computation requires $O(N^2 + NS)$ space, and at most $O(BNS + BN^2 + N^2 S)$ time.

\section{Experimental setup} \label{sec:experiments}

In order to validate our method, we performed experiments on various datasets in the computer vision domain. We evaluated both the computing time required by Hebbian algorithms, with and without the FastHebb optimization, and their performance in sample efficiency scenarios, also making comparisons with backprop-based learning. In the following, we describe the details of our experiments and comparisons, discussing the network architecture and the training procedure.

\subsection{Datasets and sample efficiency regimes}
The datasets that we considered for our experiments are CIFAR10 \cite{cifar}, CIFAR100, Tiny ImageNet \cite{tinyimagenet}, and ImageNet \cite{imagenet}. We performed our experiments in various regimes of label scarcity. We define an s\% \textit{sample-efficiency regime} as a scenario in which on s\% of the training set elements is assumed to be labeled. We considered 1\%, 2\%, 3\%, 4\%, 5\%, 10\%, 25\%, and 100\% sample efficiency regimes.

For each of the above regimes, we run our experiments in a semi-supervised training fashion: first, an unsupervised pre-training stage was performed, exploiting the Hebbian learning rules, using all the available training samples; this was followed by a supervised backprop-based fine-tuning stage on the labeled samples only.

\subsection{Network architecture and training}

\begin{figure}[t]
\centering
\includegraphics[width=0.8\textwidth]{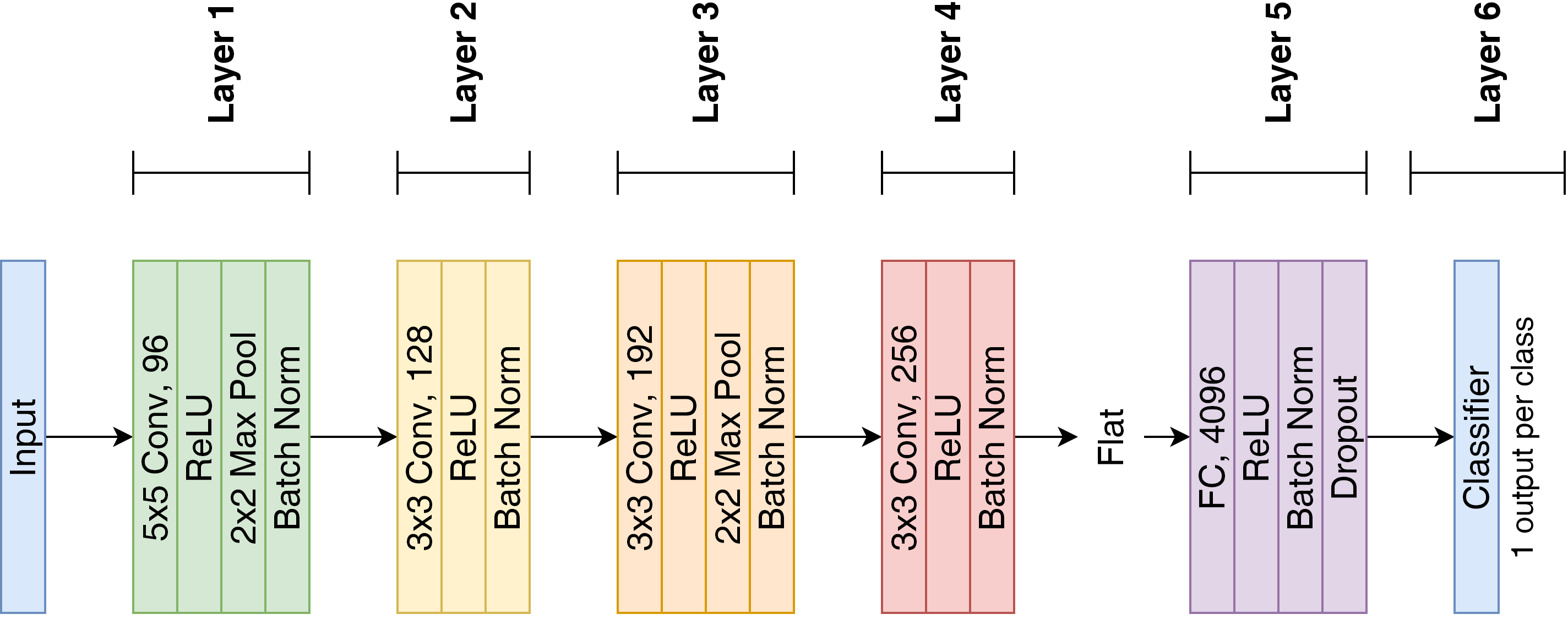}
\caption{The neural network used for the experiments.}
\label{fig:network}
\end{figure}

We considered a six layer neural network as shown in Fig. \ref{fig:network}: five deep layers plus a final linear classifier. The various layers were interleaved with other processing stages (such as ReLU nonlinearities, max-pooling, etc.), and the overall architecture was inspired by AlexNet \cite{krizhevsky2012}. 

\begin{figure}[t]
\centering
\includegraphics[width=\textwidth]{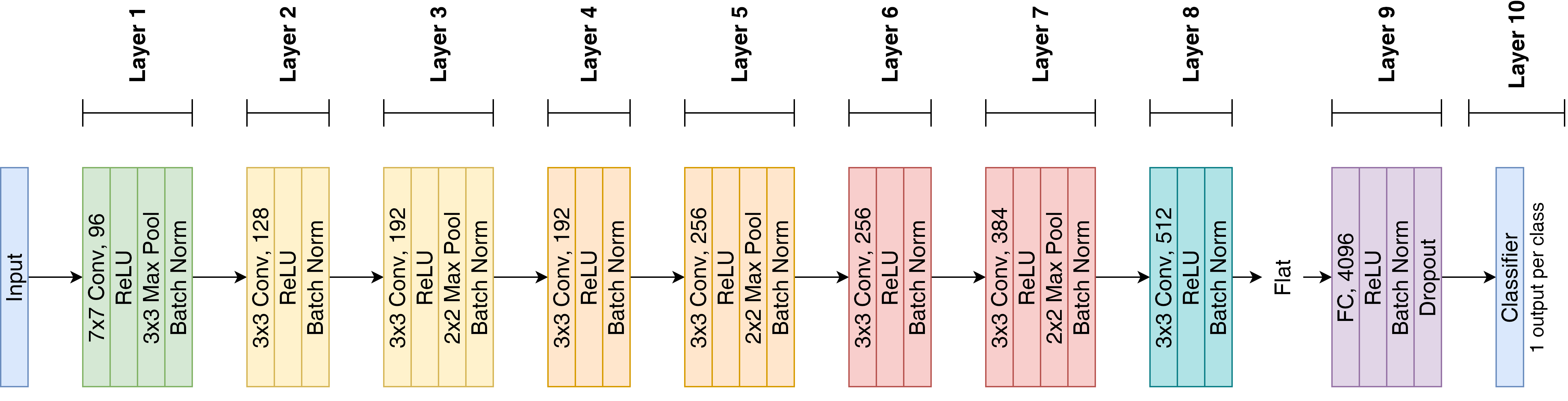}
\caption{The bigger neural network used for the experiments on ImageNet.}
\label{fig:network_10l}
\end{figure}

A similar, but bigger model was used for ImageNet classification, which is shown in Fig. \ref{fig:network_10l}.

For each sample efficiency regime, we trained the network with our semi-supervised approach in a classification task. First, we used Hebbian unsupervised pre-training rules in the internal layers. This was followed by the fine tuning stage with SGD training, involving the final classifier as well as the previous layers, in an end-to-end fashion.

For each configuration we also created a baseline for comparison. In this case, we used another popular unsupervised method, namely the Variational Auto-Encoder (VAE) \cite{kingma2013}, for the unsupervised pre-training stage. This was again followed by the supervised end-to-end fine tuning based on SGD. VAE-based semi-supervised learning was also the approach considered in \cite{kingma2014b}.

Both classification accuracy and training time were evaluated and used as metrics for comparisons.

\subsection{Details of training}

We implemented our experiments using PyTorch. All the hyperparameters mentioned below resulted from a parameter search aimed at maximizing the validation accuracy on the respective datasets, following the Coordinate Descent (CD) approach \cite{kolda2003}. 

Training was performed in 20 epochs using mini-batches of size 64. No more epochs were necessary, since the models had already reached convergence at that point. Networks were fed input images of size 32x32 pixels, except for the case of ImageNet, where images of size 210x210 were used.


During Hebbian training, the learning rate was set to $10^{-3}$ ($10^{-4}$ for ImageNet). No L2 regularization or dropout was used, since the learning method did not present overfitting issues.

For VAE training, the network backbone without the classifier acted as encoder, with an extra layer mapping the output to 256 gaussian latent variables, while a specular network branch acted as decoder.
VAE training was performed without supervision, in an end-to-end encoding-decoding task, optimizing the $\beta$-VAE Variational Lower Bound \cite{higgins2016}, with coefficient $\beta = 0.5$.

Both for VAE training and for the supervised training stage, based on SGD, the initial learning rate was set to $10^{-3}$ and kept constant for the first ten epochs, while it was halved every two epochs for the remaining ten epochs. We also used momentum coefficient $0.9$, and Nesterov correction. During supervised training, we also used dropout rate 0.5, L2 weight decay penalty coefficient set to $5 \cdot  10^{-2}$ for CIFAR10, $10^{-2}$ for CIFAR100, $5 \cdot 10^{-3}$ for Tiny ImageNet, and $1 \cdot 10^{-3}$ for ImageNet. Cross-entropy loss was used as optimization metric.

To obtain the best possible generalization, \textit{early stopping} was used in each training session, i.e. we chose as final trained model the state of the network at the epoch when the highest validation accuracy was recorded.

Experiments were performed on an Ubuntu 20.4 machine, with Intel Core I7 10700K Processor, 32GB Ram, and NVidia Geforce 3060 GPU with 12GB dedicated memory. The experiments were implemented using the Pytorch package, version 1.8, and Python 3.7.

\section{Results and discussion} \label{sec:results}

In this section, the experimental results obtained with each dataset are presented and analyzed. We report the training times on each dataset, for all the approaches explored. Moreover, we report the classification accuracy in the semi-supervised task, in the various sample efficiency regimes. Experiment results from five independent iterations were averaged, and the differences between methods were tested for statistical significance with a p value of 0.05.

\subsection{Training time performance evaluation}

\begin{table}[t]
    \centering
    \caption{Training times on each dataset, for VAE, Hebbian PCA (HPCA), Hebbian PCA with FastHebb (HPCA-FH), soft-WTA (SWTA), and soft-WTA with FastHebb (SWTA-FH) methods.}
    \begin{tabular}{c|c|ccc}
    \hline
        \textbf{Dataset} & \textbf{Method} & \textbf{Epoch Duration} & \textbf{Num. Epochs} & \textbf{Total Duration} \\
        \hline
        \multirow{5}{*}{CIFAR10}
            & VAE & 14s & 17 & 3m 58s \\
            & SWTA & 4m 14s & 1 & 4m 14s \\
            & SWTA-FH & 18s & 1 & \textbf{18s} \\
            & HPCA & 6m 23s & 12 & 1h 16m 36s \\
            & HPCA-FH & 19s & 12 & 3m 48s \\
        \hline
        \multirow{5}{*}{CIFAR100}
            & VAE & 15s & 15 & 3m 45s \\
            & SWTA & 4m 16s & 1 & 4m 16s \\
            & SWTA-FH & 18s & 1 & \textbf{18s} \\
            & HPCA & 6m 25s & 7 & 44m 55s \\
            & HPCA-FH & 19s & 7 & 2m 13s \\
        \hline
        \multirow{5}{*}{Tiny ImageNet}
            & VAE & 33s & 20 & 11m \\
            & SWTA & 9m 41s & 1 & 9m 41s \\
            & SWTA-FH & 41s & 1 & \textbf{41s} \\
            & HPCA & 14m 20s & 14 & 3h 20m 40s \\
            & HPCA-FH & 43s & 14 & 10m 2s \\
        \hline
        \multirow{5}{*}{ImageNet}
            & VAE & 2h 59m 19s & 16 & 47h 49m 4s \\
            & SWTA & 105h 13m 24s & 3 & 315h 40m 12s \\
            & SWTA-FH & 3h 38m 6s & 3 & \textbf{10h 54m 18s} \\
            & HPCA & 155h 41m 39s & 3 & 467h 4m 57s \\
            & HPCA-FH & 3h 39m 18s & 3 & 10h 57m 54s \\
        \hline
    \end{tabular}
    \label{tab:traintimes}
\end{table}

Table \ref{tab:traintimes} shows the training time measured on the various datasets, for each of the considered approaches. We measured the average epoch duration, the total number of training epochs required by each method, and the total training time. The number of epochs is counted by considering the training over when the network performance stops improving. The reported number of epochs refers to the pre-training phase only, and not to the successive fine-tuning, as we observed no statistically significant difference in the duration of the latter phase for different pre-training methods. Training time of FastHebb methods are compared to the previous respective best known solutions for Hebbian learning, that were also based on GPU \cite{lagani2022a}.

We can see that, in terms of total training time, Hebbian methods are almost five times faster than VAE on ImageNet.
Among the Hebbian approaches, soft-WTA is faster, thanks to its lower time complexity.
Most importantly, as shown form the ImageNet performance results, thanks to the novel optimization, FastHebb algorithms scale gracefully also to large scale datasets.

\subsection{Semi-supervised, sample efficiency scenario}

\begin{table}[t]
    \centering
    \caption{Accuracy results on each dataset (top-1 for CIFAR10, and top-5 for the other datasets, since they have many more classes), for the various approaches explored.}
    \begin{tabular}{c|c|cccc}
    \hline
        \textbf{Regime} & \textbf{Method} & \textbf{CIFAR10} & \textbf{CIFAR100} & \textbf{Tiny ImageNet} & \textbf{ImageNet} \\
        \hline
        \multirow{3}{*}{1\%}
            & VAE & 22.54  & 12.28  & 5.55  & 2.72 \\
            & SWTA & 30.23  & 15.30  & 6.20  & 6.69 \\
            & HPCA & \textbf{39.75}  & \textbf{22.63}  &  \textbf{11.38}  & \textbf{8.65} \\
        \hline
        \multirow{3}{*}{2\%}
            & VAE & 26.78  & 15.25  & 6.74  & 6.14 \\
            & SWTA & 36.59  & 20.76  & 8.56  & 11.52 \\
            & HPCA & \textbf{45.51}  & \textbf{30.83}  & \textbf{15.71}  & \textbf{13.64} \\
        \hline
        \multirow{3}{*}{3\%}
            & VAE & 29.00  & 16.44  & 7.74  & 15.35 \\
            & SWTA & 41.54  & 23.69  & 10.26  & 15.67 \\
            & HPCA & \textbf{48.80}  & \textbf{35.04}  & \textbf{18.23}  & \textbf{17.28} \\
        \hline
        \multirow{3}{*}{4\%}
            & VAE & 31.15  & 17.89  & 8.45  & \textbf{23.97} \\
            & SWTA & 45.31  & 26.91  & 11.52  & 19.95 \\
            & HPCA & \textbf{51.28}  & \textbf{38.89}  & \textbf{20.55}  & 20.39 \\
        \hline
        \multirow{3}{*}{5\%}
            & VAE & 32.75  & 18.48  & 9.29  & \textbf{29.04} \\
            & SWTA & 48.35  & 29.57  & 12.55  & 24.87 \\
            & HPCA & \textbf{52.20}  & \textbf{41.42}  & \textbf{22.46}  & 23.28 \\
        \hline
        \multirow{3}{*}{10\%}
            & VAE & 45.67  & 23.80  & 13.51  & \textbf{43.73} \\
            & SWTA & \textbf{58.00}  & 38.26  & 16.70  & 41.54 \\
            & HPCA & 57.35  & \textbf{48.93}  & \textbf{28.13}  & 34.27 \\
        \hline
        \multirow{3}{*}{25\%}
            & VAE & 68.70  & 52.59  & \textbf{37.89}  & \textbf{61.33} \\
            & SWTA & \textbf{69.85}  & 56.26  & 24.96  & 59.34 \\
            & HPCA & 64.77  & \textbf{58.70}  & 37.10  & 56.92 \\
        \hline
        \multirow{3}{*}{100\%}
            & VAE & 85.23  & \textbf{79.97}  & \textbf{60.23} & 76.84  \\
            & SWTA & \textbf{85.37}  & 79.80  & 54.94  & 76.10 \\
            & HPCA & 84.38  & 74.42  & 53.96  & \textbf{77.28} \\
        \hline
    \end{tabular}
    \label{tab:accuracy}
\end{table}

Table \ref{tab:accuracy} shows the classification accuracy results obtained on the various dataset, for each of the considered approaches. Top-1 accuracy was used for CIFAR10, and top-5 for all the other datasets, since they have many more classes. Note that, in this case, we show the results for HPCA and soft-WTA, but these are the same with or without the FastHebb optimization. In fact, the optimization does not change the update rule itself.

We can observe that Hebbian approaches perform better than VAE in sample efficiency regimes with very scarce label availability, below 4-5\%. In particular, we can observe performance improvements of HPCA of almost 20\% in the 5\% regime for the CIFAR10 dataset. On the other hand VAE-based pre-training only improves when the available number of labeled training samples for the successive supervised fine-tuning phase becomes larger. 
When scaling up to ImageNet dataset, we still have a slight advantage of Hebbian methods in scarce data regimes (from 2 to 6\%, depending on the regime). However, when higher regimes are considered, the performance of Hebbian pre-training is slightly lower than VAE, but this is compensated, as shown before, by a significant advantage in terms of training time.

\section{Conclusions and future work} \label{sec:conclusions}

We have shown how the FastHebb approach can be leveraged to optimize running times of Hebbian learning algorithms for DNN training. Thanks to this optimization, we were able to scale Hebbian learning experiments to ImageNet level. To the best of our knowledge, this is the first solution able to bring Hebbian learning to such scale. Experiments in semi-supervised scenarios show the efficacy of Hebbian approaches for unsupervised network pre-training, compared to backprop-based VAE pre-training, both in terms of classification accuracy and training time, especially in sample efficiency scenarios where the labeled data for supervised fine tuning are scarce (less than 4-5\% of the overall available data). 

As possible future work directions, we suggest to perform further studies of FastHebb on other large-scale application scenarios, such as Content Based Image Retrieval (CBIR) to evaluate the quality of deep features extracted by this method. Preliminary work in this direction is promising \cite{lagani2022b}. 
Moreover, further Hebbian rules can also be derived, for example from Independent Component Analysis (ICA) \cite{hyvarinen} and sparse coding \cite{olshausen1996a,olshausen1996b,rozell2008}. 
Finally, in the context of semi-supervised learning, Hebbian approaches can also be combined with pseudo-labeling and consistency-based methods mentioned in Section \ref{sec:rel_work} \cite{iscen2019, sellars2021}.

%
%
%
\bibliographystyle{splncs04}
\bibliography{references}

\end{document}